\documentclass{article}

\usepackage{microtype}
\usepackage{graphicx}
\usepackage{subfigure}
\usepackage{array}
\usepackage{float}
\usepackage{pifont}
\usepackage{booktabs} 

\usepackage{hyperref}

\usepackage[accepted]{icml2023}

\usepackage{amsmath}
\usepackage{amssymb}
\usepackage{mathtools}
\usepackage{amsthm}
\usepackage{xurl}

\usepackage[capitalize,noabbrev]{cleveref}

\usepackage{alexdefs}

\theoremstyle{plain}

\theoremstyle{definition}

\theoremstyle{remark}

\usepackage[textsize=tiny]{todonotes}

\icmltitlerunning{Benchmarking Bayesian Causal Discovery Methods for Downstream Treatment Effect Estimation}

\begin{document}

\twocolumn[
\icmltitle{Benchmarking Bayesian Causal Discovery Methods\\for Downstream Treatment Effect Estimation}




\begin{icmlauthorlist}
\icmlauthor{Chris Chinenye Emezue \textsuperscript{\dag}}{ce,mila}
\icmlauthor{Alexandre Drouin}{mila,ad}
\icmlauthor{Tristan Deleu}{mila,udem}
\icmlauthor{Stefan Bauer}{ce,helm}
\icmlauthor{Yoshua Bengio}{mila,udem,cifar,cifar2}
\end{icmlauthorlist}

\icmlaffiliation{ce}{Technical University of Munich, Munich, Germany}
\icmlaffiliation{ad}{ServiceNow Research, Montreal, Canada}
\icmlaffiliation{udem}{Université de Montréal, Montreal, Canada}
\icmlaffiliation{mila}{Mila - Quebec AI Institute, Montreal, Canada}
\icmlaffiliation{cifar}{CIFAR AI Chair}
\icmlaffiliation{cifar2}{CIFAR Senior Fellow}
\icmlaffiliation{helm}{Helmholtz AI}

\icmlcorrespondingauthor{Chris Chinenye Emezue}{chris.emezue@gmail.com}

\icmlkeywords{gflownets, treatment effect, causal discovery, dag-gflownet, causal inference}

\vskip 0.3in
]



\printAffiliationsAndNotice{\dag Work done as a visiting research student at Mila.}  

\begin{abstract}
The practical utility of causality in decision-making is widespread and brought about by the intertwining of causal discovery and causal inference. Nevertheless, a notable gap exists in the evaluation of causal discovery methods, where insufficient emphasis is placed on downstream inference. To address this gap, we evaluate seven established baseline causal discovery methods including a newly proposed method based on GFlowNets, on the downstream task of treatment effect estimation. Through the implementation of a distribution-level evaluation, we offer valuable and unique insights into the efficacy of these causal discovery methods for treatment effect estimation, considering both synthetic and real-world scenarios, as well as low-data scenarios. The results of our study demonstrate that some of the algorithms studied are able to effectively capture a wide range of useful and diverse ATE modes, while some tend to learn many low-probability modes which impacts the (unrelaxed) recall and precision.

\end{abstract}

\section{Introduction}


Causal inference has a wide variety of real-world applications in domains such as healthcare \cite{pmlr-v89-tu19a,huang2019diagnosis,bica2020estimating,deci} , marketing \cite{zhangHealth,battocchi2021estimating,sanchez2022causal}, political science, and online advertising \cite{waisman2019online,causalnlp}. Treatment effect estimation, the process of estimating the effect or impact of a treatment on an outcome in the presence of other covariates as potential confounders (and mediators), is a fundamental problem in causal inference that has received widespread interest for decades \cite{chu2023causal}. 

The existing powerful methods for treatment effect estimation from data require a complete (or partial) a priori knowledge of the causal graph \cite{judeaPearlCausalInference,bookofwhy}. When the graph is unknown, this requires solving a problem of \emph{causal structure learning}, also known as \emph{causal discovery}. Structure learning involves learning a graph (typically characterized by a directed acyclic graph or DAG for short) that best describes the dependence structure in a given data set \cite{doi:10.1146/annurev-statistics-060116-053803,heinze-deml2017causal,glymour2019review}. In this approach, structure learning is required to learn a causal graph, which can then be applied to \textit{infer} the influence of treatments on the outcomes of interest~\cite{judeaPearlCausalInference,naser2022causality}. It should be noted that the actual causal graph can only be inferred up to its Markov Equivalence class (MEC), and the available observational data does not offer any means of further differentiation \cite{maathuis2008estimating,pearl2009}. 
Learning a single graph has been shown to lead to poor predictions in a downstream causal inference task \cite{doi:10.1080/03610929508831616,bcdnets,tigas2022interventions}.

Instead of learning a single causal graph, the problem of structure learning can be tackled from a Bayesian perspective where we learn a posterior over the causal graphs. This has the unique advantage of accounting for epistemic uncertainty over the causal graphs in the MEC, thereby leading to a more enriching predictive performance in a downstream causal inference task. However, learning such a posterior over the causal graphs is plagued by challenges. One major issue is the combinatorially large sample space of causal graphs. The second major challenge is related to MCMC mode-mixing \cite{10.2307/20061160,bengio2012better}: the mode-mixing problem occurs when the chances of going from one mode to a neighboring one may become exponentially small and require exponentially long chains, if the modes are separated by a long sequence of low-probability configurations. Therefore by using MCMC, there is an important set of distributions for which finite chains are unlikely to provide enough diversity of the modes of the distribution \cite{bengio2021gflownet}. While there are a number of existing causal discovery methods (both Bayesian and non-Bayesian), our benchmark study centers on DAG-GFlowNet \cite{daggfn}, which is a unique method that leverages a novel class of probabilistic models called \textit{Generative Flow Networks} ~\cite{bengio2021flow,bengio2021gflownet} to approximate the posterior distribution over causal graphs. 

Although causal inference is an inherent downstream application of causal discovery, most causal discovery evaluation methods are not aligned with causal inference because these two fields are typically studied independently \cite{deci}.
For example, many causal discovery evaluation methods use the structural hamming distance (SHD) which compares the learned causal DAG (or the samples from the posterior distribution of DAGs in Bayesian structure learning) to the true DAG of the data generating process. Measuring the proximity of the learned DAGs, however, does not reveal much about their actual performance in treatment effect estimation given a treatment and outcome variable of interest, which is a predominantly downstream evaluation.

In this work, we set out to benchmark causal discovery methods for the downstream task of treatment effect estimation, specifically the average treatment effect. As an extension to the DAG-GFlowNet, we offer insights on the application of GFlowNets to average treatment effect estimation, by comparing it with six other baseline methods for causal discovery.

\section{Background}
We provide a detailed background, in \autoref{bg}, on some of the key concepts used in this paper: Bayesian network, interventional distribution, Bayesian causal discovery, average treatment effect and our structure learning baselines. The structure learning baselines employed in our study follow \citet{daggfn}. In addition to DAG-GFlowNet \cite{daggfn}, we leveraged six baseline causal discovery algorithms: PC \cite{pc}, GES \cite{ges}, MC3 \cite{mc3}, BCDNets \cite{bcdnets}, Gadget \cite{gadget}, and DiBS \cite{dibs}. Due to space restrictions, we move our explanation of the causal discovery methods to Section \ref{cdd} in the Appendix.

\section{Experimental setup}

\begin{figure*}[!t]
\includegraphics[width=\textwidth]{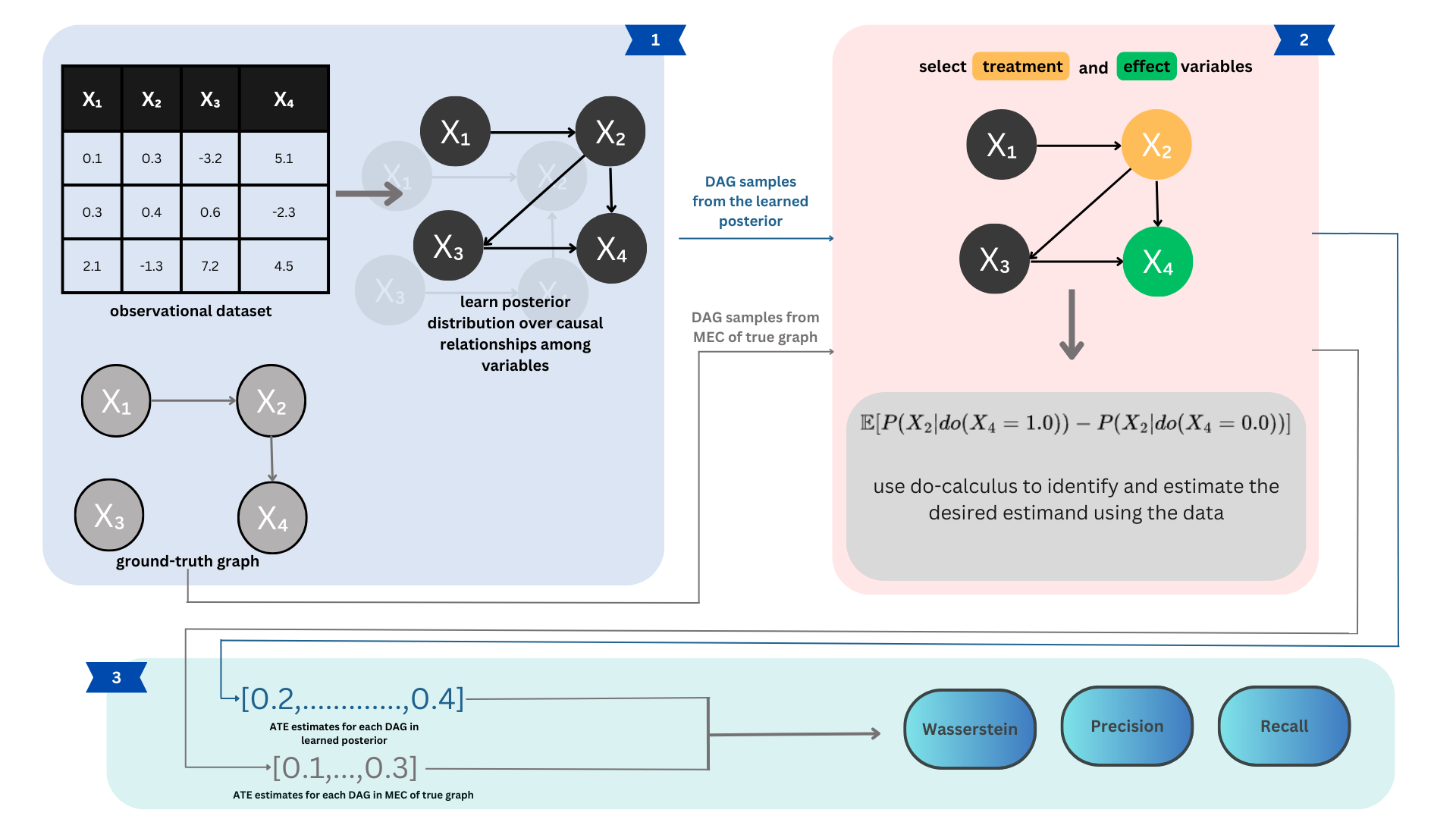} %
\centering
\caption{Illustration of our experimental pipeline using four random variables. We start with (Bayesian) causal discovery which involves learning a posterior over DAGs that best explain the training dataset. The next step is average treatment effect estimation (ATE). For each pair of distinct variables, the second stage involves estimating its ATE for each DAG in the posterior. Furthermore, from the true graph, we enumerate the DAGs in its Markov equivalence  class (MEC) and find the ATE estimates for all of them. This allows us to perform our evaluation which involves comparing the ATE distributions between the true graph MEC and learned posterior distribution of DAGs. For clarity, this illustration has been given using only one baseline and one seed. For our experiments, we worked with 6 baselines and 26 seeds for each baseline.}
\label{our-workflow}
\end{figure*}


Figure \ref{our-workflow} provides an illustrative overview of our experimental pipeline. The initial step involves Bayesian causal discovery, where, as discussed in Section \ref{bayesian-cd}, the objective is to learn a posterior distribution of the directed acyclic graphs (DAGs) that provide the most plausible explanations for the training dataset. The subsequent stage involves the estimation of the average treatment effect (ATE). Here, the ATE for each DAG in the posterior is estimated for every pair of distinct variables. In addition, the DAGs within the Markov equivalence class (MEC) of the true graph are enumerated and used to calculate the ATE estimates for each of them. The evaluation process, in stage 3, then involves a comparison of the average treatment effect (ATE) distributions between the true graph Markov equivalence class (MEC) and the learned posterior distribution of DAGs. 

For our experiments on synthetic data, we worked with 6 baselines in total and 26 seeds for each baseline. Each seed corresponds to a causal discovery experiment with a randomly sampled truth graph and observational data.

\subsection{Causal discovery experiments}
Following \citet{daggfn}, we performed causal discovery experiments on synthetic and real-world scenarios. For PC and GES we implement bootstrapping to achieve DAG posterior samples. 

\paragraph{Analysis on synthetic data:}
Following \citet{daggfn}, we performed experimental analyses using synthetic graphs and simulated data. We sampled synthetic data from linear Gaussian Bayesian networks with randomly generated structures. We experimented with Bayesian networks of size $d=20$ variables and considered two different sample sizes of $n=20$ and $n=100$. A small sample size of 20 was specifically chosen to evaluate the capabilities of the causal discovery algorithms in a low-data regime. The ground-truth graphs are sampled according to an Erdos-Rényi model.


\paragraph{Analysis on flow cytometry data:}
DAG-GFlowNet was evaluated against the baselines on real-world flow cytometry data \cite{sachs} to learn protein signaling pathways. The data consists of continuous measurements of $d = 11$ phosphoproteins in individual T-cells. They used the first $n = 853$ observations and the DAG, inferred by \citet{sachs} and containing 11 nodes and 17 edges, as the dataset and ground-truth graph respectively for their causal discovery experiments. We continued with this direction in our experimental analysis and our goal was to show the downstream performance of DAG-GFlowNet on average treatment effect of the phosphoproteins in the protein signaling pathways.

\subsection{ATE experiments}

For our ATE experiments, we utilized all pairs of distinct variables: the rationale behind this was to thoroughly explore the possible treatment effects across various combinations. Therefore given $d$ random variables $\{X_{1},...,X_{d}\}$, we performed ATE evaluations on $d^2 - d$ variable pairs. To achieve this in practice, we leveraged the DoWhy package \cite{dowhy,dowhypaper}, which facilitated the implementation of the do-calculus algorithm. To ensure consistency and clarity in our results, we set the treatment values at $1.0$ and $0.0$ for all our experiments. The choice of values 1.0 and 0.0 does not relate to the existence or absence of a treatment, as is commonly used in most causal inference literature. 

Performing such a robust experiment involved a huge computation load. For example, for our baselines, each with 26 random seeds, each consisting of 1000 DAG samples from the posterior, we had to do $d*(d-1) * 1000 * 26 * 6$ ATE estimations. For the synthetic graph with 20 nodes, this leads to 57M estimations. In order to optimize the computational efficiency of our experiments, we implemented parallelism techniques. The GNU parallel computing tool \cite{Tange2011a} enabled us to distribute the computational workload across multiple processors or cores, thereby significantly reducing the overall computation time. 

\subsection{Evaluation framework}
\label{evaluation}


Our evaluation methodology goes beyond single-point ATE estimation, which is employed in standard causal inference benchmarking, by performing ATE evaluations based on posterior samples. This approach aims to provide a more comprehensive assessment of the quality of the learned posterior average treatment effect (ATE). Specifically, our evaluation pipeline involves the following metrics:

\paragraph{Wasserstein distance (WD):}
To obtain a quantitative measure of the similarity between the true ATE sample-based distrbution and that of the learned ATE, we calculate and report their Wasserstein distance \cite{ramdas2017wasserstein} using their samples\footnote{We utilize the Python implementation available \hyperlink{https://docs.scipy.org/doc/scipy/reference/generated/scipy.stats.wasserstein_distance.html}{here}.}.

\paragraph{Precision and Recall:}
We compute the precision and recall of the modes present in the learned ATE distribution and compare them to the modes in the true ATE distribution. In order to calculate the precision and recall, we first identify the unique modes for each of the true, $A_{T}$ and learned $A_{'}$ ATE samples. Then based on these set of modes, we calculate the true positive (modes from $A_{T}$ that are found in $A_{'}$) , false negative (modes from $A_{T}$ that are missed in $A_{'}$), and false positive (modes from $A_{'}$ that are not in $A_{T}$). Note that the lists $A_{T}$ and $A_{'}$ have been regrouped prior to running the evaluation (see Section \ref{regroup-ate}).

\subsection{Additional settings}

\paragraph{Enumerating the MEC of the true graph:}
In order to achieve our evaluation using our strategy (see Section \ref{evaluation}), it is necessary to not work with just one true graph. For a given ground-truth graph, we enumerate all the DAGs in its Markov equivalence class (MEC).

\paragraph{Regrouping ATE values:}
\label{regroup-ate}
The estimation of average treatment effects (ATE) through regression analysis is susceptible to generating estimates that may exhibit slight variations within numerical precision (e.g., 1.000000001 and 1). As our precision and recall metrics essentially perform `hard matches" on floating point values, it becomes crucial to consider the influence of numerical precision. In order to accomplish this objective, we group ATE values that are numerically close. More details are in \autoref{appendix:regroup}.

\section{Results \& Discussion}
The results presented in Table \ref{syn-table} illustrate the Wasserstein distance (WD), precision, and recall metrics of all baseline methods in terms of their learned ATE samples. Upon examining the Wasserstein distance, 
PC achieves the lowest Wasserstein distance, while GES attains the highest. 

\begin{table}[h!]
\caption{ATE Evaluation on synthetic data (20 variables, 20 samples). We chose a small sample size of 20 in order to evaluate the capabilities of the causal discovery algorithms in a low-data regime. We report the Wasserstein distance (WD), precision and recall of all baselines. Mean is taken over all the $20*19$ variable combination pairs and over the 26 seeds for each baseline. Standard error is taken over the 26 independent seeds. }
\label{syn-table}
\vskip 0.15in
\begin{small}
\begin{sc}
\resizebox{\columnwidth}{!}{%
\begin{tabular}{lcccr}
\toprule
 Method & WD $\downarrow$ & Precision $\uparrow$ & Recall $\uparrow$ \\
\midrule

bcdnets & 0.257 $\pm$ 0.0257 & \textbf{0.81} $\pm$ \textbf{0.0118} & 0.75 $\pm$ 0.0102 \\
bootstrap-ges & 0.328 $\pm$ 0.0309 & 0.01 $\pm$ 0.0005 & \textbf{0.98} $\pm$ \textbf{0.0025} \\
bootstrap-pc & \textbf{0.256} $\pm$ \textbf{0.0303} & 0.52 $\pm$ 0.0106 & 0.79 $\pm$ 0.0097 \\
dibs & 0.263 $\pm$ 0.0299 & 0.49 $\pm$ 0.0298 & 0.80 $\pm$ 0.0076 \\
gadget & 0.304 $\pm$ 0.0302 & 0.01 $\pm$ 0.0006 & 0.94 $\pm$ 0.0065 \\
mc3 & 0.297 $\pm$ 0.0282 & 0.01 $\pm$ 0.0007 & 0.94 $\pm$ 0.0062 \\

\hline
dag-gflownet & 0.325 $\pm$ 0.0284 & 0.01 $\pm$ 0.0005 & 0.97 $\pm$ 0.0033 \\
\bottomrule
\end{tabular}
}
\end{sc}
\end{small}
\end{table}



When focusing on precision, we observe that apart from BCDNets, all the methods seem to be performing very poorly. However all the methods attain relatively high recall scores, with the highest achieved by GES and closely followed by DAG-GFlowNet. This high recall indicates the ability of the methods to capture diverse modes within their ATE distribution. 

\begin{table}[h!]
\caption{ATE Evaluation on synthetic data (20 variables, 100 samples). We report the Wasserstein distance (WD), precision and recall of all baselines. Mean is taken over all the 380 distinct pairs and over the 26 seeds for each baseline. Standard error is taken over the 26 independent seeds.}
\label{syn-table-100}
\vskip 0.15in
\begin{center}
\begin{small}
\begin{sc}
\resizebox{\columnwidth}{!}{
\begin{tabular}{lcccr}
\toprule
 Method & WD $\downarrow$ & Precision $\uparrow$ & Recall $\uparrow$ \\
\midrule

bcdnets & \textbf{0.100} $\pm$ \textbf{0.0092} & \textbf{0.79} $\pm$ \textbf{0.0141} & 0.79 $\pm$ 0.0122 \\
bootstrap-ges & 0.165 $\pm$ 0.0211 & 0.01 $\pm$ 0.0011 & \textbf{0.99} $\pm$ \textbf{0.0039} \\
bootstrap-pc & 0.178 $\pm$ 0.0215 & 0.50 $\pm$ 0.0242 & 0.82 $\pm$ 0.0112 \\
dibs & 0.187 $\pm$ 0.0214 & 0.05 $\pm$ 0.0069 & 0.97 $\pm$ 0.0037 \\
gadget & 0.244 $\pm$ 0.0228 & 0.02 $\pm$ 0.0016 & 0.91 $\pm$ 0.0088 \\
mc3 & 0.177 $\pm$ 0.0220 & 0.06 $\pm$ 0.0064 & 0.93 $\pm$ 0.0138 \\

\hline
dag-gflownet & 0.201 $\pm$ 0.0234 & 0.03 $\pm$ 0.0051 & 0.95 $\pm$ 0.0116 \\
\bottomrule

\end{tabular}
}
\end{sc}
\end{small}
\end{center}
\vskip -0.1in
\end{table}

The WD, precision, and recall for the synthetic data experiments with 100 samples are presented in Table \ref{syn-table-100}. Given an increased number of observational samples compared to the previous table, it is anticipated that the task of causal discovery will be simpler. This is evidenced in the lower WD scores compared to Table \ref{syn-table}. In a manner similar to the scenario involving 20 samples, it is observed that the methods, with the exception of BCDNets, exhibit a considerably low precision score, while concurrently displaying high recall values.

\begin{table}[h!]
\caption{ATE Evaluation on the Sachs dataset \cite{sachs}. We report the Wasserstein distance (WD), precision and recall of all baselines. Mean and standard deviation are taken over all the $110$ distinct pairs. Only one seed was used in this experiment, so standard deviation was computed instead of standard error.}
\label{sachs=table}
\vskip 0.15in
\begin{center}
\begin{small}
\begin{sc}
\resizebox{\columnwidth}{!}{%
\begin{tabular}{lccc}
\toprule
 Method& WD $\downarrow$ & Precision $\uparrow$ & Recall $\uparrow$ \\
\midrule


bcdnets & 0.040 $\pm$ 0.1086 & \textbf{0.97} $\pm$ \textbf{0.1489} & 0.65 $\pm$ 0.3609 \\
bootstrap-ges & \textbf{0.037} $\pm$ \textbf{0.0992} & 0.13 $\pm$ 0.1291 & 0.96 $\pm$ 0.0941 \\
bootstrap-pc & 0.038 $\pm$ 0.1035 & 0.83 $\pm$ 0.2391 & 0.82 $\pm$ 0.2700 \\
dibs & \textbf{0.037} $\pm$ \textbf{0.1011} & 0.17 $\pm$ 0.1420 & 0.94 $\pm$ 0.1331 \\
gadget & 0.038 $\pm$ 0.1033 & 0.42 $\pm$ 0.2560 & 0.92 $\pm$ 0.1444 \\
mc3 & 0.041 $\pm$ 0.1115 & 0.46 $\pm$ 0.3273 & 0.86 $\pm$ 0.2359 \\

\hline
dag-gflownet & 0.039 $\pm$ 0.1074 & 0.14 $\pm$ 0.1918 & \textbf{0.98} $\pm$ \textbf{0.0819} \\
\bottomrule

\end{tabular}
}
\end{sc}
\end{small}
\end{center}
\vskip -0.1in
\end{table}







\begin{figure*}[htb!]
    \centering
    \includegraphics[width=\textwidth]{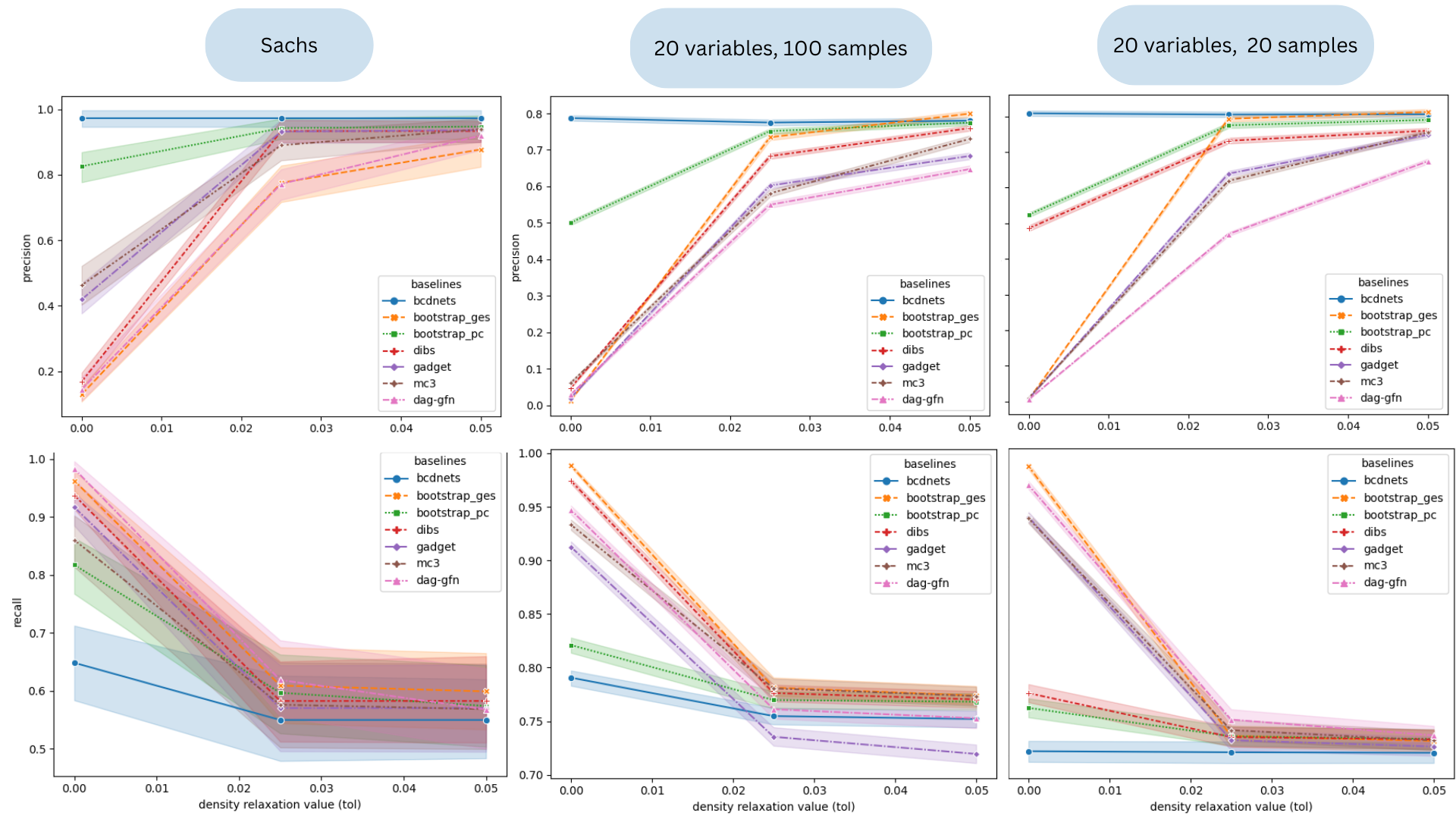}
    \caption{Lineplot of the average and standard deviation of the precision (top) and recall (bottom) for each of the baseline algorithms. The lineplots are visualized for each dataset used in our experiment, denoted by the columns.}
    \label{fig:ate-relax-all}
\end{figure*}

Table \ref{sachs=table} presents the evaluation results of the analysis on flow cytometry using the Sachs dataset. Overall, all methods demonstrate comparable performance in terms of the Wasserstein distance: the range of the WD is 0.004, unlike in Table \ref{syn-table} which is 0.072 or Table \ref{syn-table-100} which is 0.144. When considering precision, BCDNets and PC outperform DAG-GFlowNet, which exhibits lower performance. Notably, DAG-GFlowNet achieves the highest recall, indicating its ability to learn samples from diverse modes within the true ATE distribution.

\subsection{Filtering Low-Probability Modes}

In all our evaluations (Tables \ref{syn-table}, \ref{syn-table-100}, \ref{sachs=table}), we witness a trend of DAG-GFlowNet and other methods exhibiting very low precision scores. 


\begin{figure}[h!]

    \centering
    \includegraphics[width=0.5\textwidth]{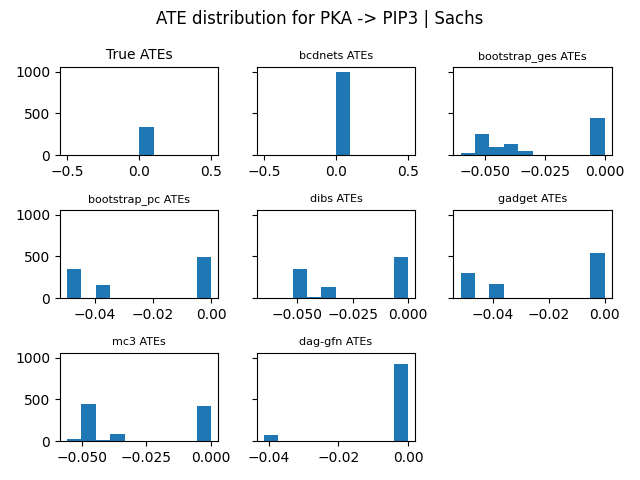}
    \caption{Comparison of ATE distribution of the methods using flow cytometry data. We focus on the treatment variable $PKA$ and the outcome variable $PIP3$ from the Sachs dataset. We see that while DAG-GFlowNet correctly captures all modes of the true distribution (just like BCDNets), it additionally learns values at 0.04 albeit with a very low probability mass.}
    \label{fig:ate-histogram-pka-pip3}
\end{figure}

In Figure \ref{fig:ate-histogram-pka-pip3} we observe that DAG-GFlowNet (and other baselines like GES, DiBs) tends to learn new modes, but those modes have a very low probability in the estimated distribution. In our current evaluation framework however, we include all values in the list that have non-zero densities, which leads to unfair penalization of methods that exhibit multimodal diversity. Consequently, these methods receive disproportionately low precision values. However, when we apply a filtering approach that removes the low-probability modes before calculating the metrics, a more insightful narrative emerges for these methods, as shown in Figure \ref{fig:ate-relax-all}. In particular, we notice a significant increase in precision for all the methods that initially exhibited very low precision values (in Tables \ref{syn-table}, \ref{syn-table-100}, and \ref{sachs=table}), when we apply a density relaxation tolerance of 0.05 (i.e for any list of ATEs, we only consider ATE values that have a mass of at least 0.05). This trend is consistent across all the experimental settings (100 samples, 20 samples, Sachs dataset). 



\section{Conclusion}

In conclusion, the practical importance of causality in decision-making is widely acknowledged, and the interplay between causal discovery and inference is evident. In order to bridge the gap in the evaluation of causal discovery methods, where limited attention is given to downstream inference tasks, we conducted a comprehensive evaluation that assessed seven established baseline causal discovery methods including a novel approach utilizing GFlowNets. By incorporating a Bayesian perspective in our evaluation, we offer a unique form of distribution-level insights, into their effectiveness for downstream treatment effect estimation.



\bibliography{refs}
\bibliographystyle{icml2023}

\newpage
\appendix
\onecolumn

\section{Related Work}
\label{relatedWork}
\paragraph{Benchmarking methods:}
Benchmarks have played a crucial role in advancing entire research fields, for instance computer vision with the introduction of ImageNet \cite{NIPS2012c399862d}. When it comes to causal discovery, benchmarks usually come in the form of research surveys \cite{10.1145/3527154}, benchmark datasets \cite{beinlich89alarm,chevalley2022causalbench,menegozzo2022cipcadbench}, learning environments \cite{Runge2019inferring,runge2017detecting,ahmed2020causalworld}, and software packages or platforms \cite{cdt, scutari2014bayesian,kalainathan2019causal}. However these methods only evaluate the closeness of the causal DAG, or the samples from the posterior distribution of DAGs in Bayesian structure learning, from various causal discovery methods to the ground-truth DAG. Measuring the proximity of the learned DAGs, however, does not reveal much about their actual performance in treatment effect estimation given a treatment and outcome variable of interest, which is a predominantly downstream evaluation.



In causal inference, datasets \cite{MacDorman1998-az,hahn2019atlantic}, frameworks \cite{shimoni2018benchmarking,ehudKaravani20181163587,neal2020realcause}, and software packages \cite{dowhy,econml} provide valuable tools for predicting the causal effects of treatments on outcomes. Causal inference plays a crucial role in decision-making and finds numerous practical applications in various domains such as healthcare, advertising, and decision-making processes. This implies that causal inference has a more downstream impact. In causal inference, the graph represents the structure of the joint distribution of variables, which is then leveraged to identify the causal estimand.

Therefore, the evaluation of causal discovery methods on downstream causal inference tasks provides more practical insights into the effectiveness and practicality of causal methods within real-world scenarios. Typically, the fields of causal discovery and inference are approached separately, resulting in limited intertwined evaluation methods. This is the aspect that distinguishes our work. Similar approaches can be found in studies that jointly integrate causal discovery and inference in an end-to-end manner, such as the notable example of DECI \cite{deci}. However, our work differs in two key aspects: firstly, we employ the novel GFlowNets for causal inference, increasing our span and secondly, we specifically focus on linear noise structural equation models, whereas DECI addresses the problem of end-to-end causal inference in non-linear additive noise structural equation models (SEM).

\section{Background}
\label{bg}
We offer a detailed background, in this section, on some of the key concepts used in this paper.

\paragraph{Bayesian network:} A (causal) Bayesian network \cite{10.5555/1795555,pearl2009} is a probabilistic model over $d$ random variables $\{X_{1},...,X_{d}\}$, whose joint probability distribution factorizes according to a DAG G (whose edges express causal dependencies) as:
\begin{equation}
    P(X_{1},...,X_{d}) = \prod_{k=1}^{d} P(X_{k} | Pa_{G}(X_k)),
\end{equation}
where $Pa_{G}(X)$ is the set of parents of the node $X$, i.e the nodes with an edge onto $X$ in G, interpreted as the direct causes of $X$.

\paragraph{Interventional distribution:} Given a random variable $X_{k}$, a (hard) intervention on $X_{k}$, denoted by $do(X_{k} = a)$, is obtained by replacing the conditional probability distribution (CPD) $P(X_{k} | Pa_{G}(X_{k}))$ with a Dirac distribution $\delta_{X_{k} = a}$ which forces $X_{k}$ to take on the value of $a$. Note that intervening on a variable, in a graphical sense, results in a mutilated graph where all incoming edges to the node corresponding to that variable are removed \cite{judeaPearlCausalInference}. 

\subsection{(Bayesian) Causal discovery}
\label{bayesian-cd}

Given a dataset $D \eqdef \left\{\xb^{(i)}\right\}_{i = 1}^n$ of $n$ observations, such that $\xb^{(j)} \sim P(X_{1},...,X_{d})$,
the goal of structure learning is to learn the DAG $G$ corresponding to the causal Bayesian network that best models $D$. It is important to note that $D$ could be observational samples or interventional data samples (got from performing hard or soft interventions). In a Bayesian structure learning setting, the task is to approximate the posterior distribution $P(G | D)$ over Bayesian networks that model these observations. A distribution over the DAGs allows quantifying the epistemic uncertainty and the degree of confidence in any given Bayesian network model, which is especially useful when the amount of data to learn from is small \cite{dibs,muller2021transformers}. 

\subsection{Average treatment effect (ATE) estimation}
The average treatment effect (ATE) is a quantity that allows us to estimate the \textit{impact} of a treatment variable on an outcome variable. Given $X_T$ and $X_Y$, our treatment and effect variables of interest respectively, the ATE on targets $X_Y$ for treatment $X_T = a$ given a reference $X_T = b$ is given by \cite{judeaPearlCausalInference}:
$$
ATE(a,b) = \mathbb{E}[X_Y|do(X_T =b)] - \mathbb{E}[X_Y|do(X_T =  a)]
$$

In practice, this causal inference is broken down into two steps: identification and estimation. Identification deals with converting the causal estimand $P(X_Y|do(X_T =b)$ into a statistical estimand that can be estimated using the dataset $D$. Some identification methods include the back-door criterion, front-door criterion \cite{judeaPearlCausalInference}, instrumental variables \cite{NBERt0136} and mediation. Causal estimation then computes the identified statistical estimand from the data set using a range of statistical methods. The do-calculus algorithm \cite{doCalculus} provides a powerful, systematic, programmable framework for the identification and estimation of the causal estimand.

\subsection{Causal discovery baseline algorithms}
\label{cdd}

In Table \ref{cd-baselines} we briefly describe the structure learning algorithms we use in this work. The structure learning baselines employed in our study follow those utilized by \citet{daggfn}. For PC and GES we implement bootstrapping to achieve DAG posterior samples. 
\begin{table}[H]
\caption{Details of baseline algorithms. We document the underlying algorithm behind each baseline, alongside an indication of whether there is a guarantee that the learned graph is a DAG. }
\label{cd-baselines}
\vskip 0.15in
\begin{center}
\begin{small}
\resizebox{0.6\columnwidth}{!}{
\begin{tabular}{lcccr}
\toprule
Baseline & Underlying algorithm& DAG Support\\
\midrule
PC & constraint-based & \ding{52}   \\
GES & score-based & \ding{52} \\
MC3 & MCMC& \ding{52}  \\
Gadget & MCMC & \ding{52} \\
DiBS & variational inference & \ding{55} \\
BCDNets & variational inference & \ding{52}\\
\hline
DAG-GFlowNet & GFlowNets & \ding{52}  \\
\bottomrule

\end{tabular}
}
\end{small}
\end{center}
\vskip -0.1in
\end{table}

\paragraph{DAG-GFlowNet:}
DAG-GFlowNet \cite{daggfn} employs GFlowNets \cite{bengio2021flow,bengio2021gflownet} as a substitute for MCMC in order to estimate the posterior distribution of Bayesian network structures, based on a set of observed data. An overview of GFlowNets is presented in Section \ref{gfn} of the Appendix. The process of creating a sample DAG from an approximate distribution is considered a sequential decision task. This involves constructing the graph incrementally, one edge at a time, by utilizing transition probabilities that have been learned by a GFlowNet. We refer the reader to \citet{daggfn} for a comprehensive study of DAG-GFlowNet.

\paragraph{DiBS:}
The DiBS framework \cite{dibs} is an approach to Bayesian structure learning that is fully differentiable. It operates within the continuous space of a latent probabilistic graph representation. In contrast to prior research, the DiBS method does not rely on a specific format for the local conditional distributions. Additionally, it enables the simultaneous estimation of the graph structure and the parameters of the conditional distributions. 

\paragraph{MC3:}
In the MC3 algorithm (also known as structured MCMC) \cite{mc3}, the authors present a hierarchical Bayesian approach to structure learning that leverages a prior over the classes of variables using nonparametric block-structured
priors over Bayes net graph structures. This approach relies heavily on the assumption that variables come in one or more classes and that the prior probability of an edge existing between two variables is a function only of their classes \cite{mc3}.
\paragraph{GES:}
The Greedy Equivalence Search (GES) algorithm \cite{ges} is a score-based method for causal discovery that has been in use for a considerable amount of time. It operates by performing a greedy search across the set of equivalence classes of DAGs. The representation of each search state is accomplished through a completed partially directed acyclic graph (CPDAG), which includes operators for the insertion and deletion of edges. These operators enable the addition or removal of a single edge, respectively \cite{hasan2023survey}.


\paragraph{PC:}
The Peter-Clark (PC) algorithm \cite{pc} is a prominent constraint-based method for causal discovery. It leverages conditional independence (CI) tests to infer the underlying causal structure. The algorithm yields a completed partially directed acyclic graph (CPDAG) that represents the relationships between variables. It follows a three-step process: 1) identifying the skeleton of the graph, 2) determining v-structures or colliders $(X \longrightarrow Y \longleftarrow Z)$ based on d-separation, and 3) propagating edge orientations. Initially, the algorithm creates a fully connected undirected graph using all variables in the dataset. It then eliminates edges that are unconditionally or conditionally independent (skeleton detection), identifies and orients v-structures using the d-separation set, and finally orients the remaining edges while ensuring the absence of new v-structures and cycles. The PC algorithm relies on the assumptions of acyclicity, causal faithfulness, and causal sufficiency.
\paragraph{BCDNets:}
BCDNets \cite{bcdnets} is another variational inference framework like DiBS. In their work they focus on estimating a distribution over DAGs characterizing a linear-Gaussian SEM and propose techniques to scale to high dimensions, such as using deep neural networks to model a variational family of factorized posterior distributions over the SEM parameters (including the edge weights and noise variance), and a horseshoe prior \cite{pmlr-v5-carvalho09a} on the edge weights, which promotes sparsity.

\paragraph{Gadget:}
Gadget \cite{gadget} is based on MCMC: sampling DAGs by simulating a Markov chain whose stationary distribution is the posterior distribution. However, to enhance the mixing of the chain, and reduce the space and time requirements, they build a Markov chain on the smaller space of ordered partitions of the node set, each state being associated with multiple DAGs.


\section{Generative Flow Networks (GFlowNets)}
\label{gfn}

The Generative Flow Networks \cite{bengio2021flow,bengio2021gflownet}, also known as GFlowNets, are a type of inference models that have a broad range of applications. GFlowNets are capable of generating samples with a probability that is proportional to a given reward function. The GFlowNets have been extensively studied and discussed in research papers such as \citet{bengio2021flow} and \citet{bengio2021gflownet}. The models facilitate the process of selecting a varied pool of potential candidates, while adhering to a training objective that ensures a nearly proportional sampling based on a specified reward function. GFlowNets are characterized unique training objectives like the \textit{flow-matching condition} \cite{bengio2021flow}, \textit{the detailed balance condition} \cite{bengio2021gflownet}, etc, through which a policy is learned. Through the training objectives, this policy is designed to ensure that the probability $P_{T}(s)$ of sampling an object $s$ is roughly proportional to the value $R(s)$ of a specified reward function applied to that object. The GFlowNets technique is designed to reduce the computational burden of MCMC methods by performing the necessary work in a single generative pass that has been trained for this purpose.

GFlowNets are well-suited for modeling and sampling from distributions over sets and graphs, as well as estimating free energies and marginal distributions \cite{jain2022biological,zhang2022generative}. They excel in problem scenarios with specific characteristics \cite{jain2023gflownets}: (1) the ability to define or learn a non-negative or non-marginalized reward function that determines the distribution to sample from, (2) the presence of a highly multi modal reward function, showcasing GFlowNets' strength in generating diverse samples, and (3) the benefit of sequential sampling, where compositional structure can be leveraged for sequential generation.

Since its inception, GFlowNets have exhibited promising results in diverse domains such as discrete probabilistic modeling \cite{zhang2022generative}, molecular design \cite{bengio2021flow,jain2022biological}, and causal discovery \cite{daggfn}. The aim of our research is to provide significant findings on the feasibility of employing GFlowNets for causal inference.

\section{Regrouping ATE values}
\label{appendix:regroup}
The estimation of average treatment effects (ATE) through regression analysis is susceptible to generating estimates that may exhibit slight variations within numerical precision (e.g., 1.000000001 and 1). As our precision and recall metrics essentially perform `hard matches" on floating point values, it becomes crucial to consider the influence of numerical precision. In order to accomplish this objective, we group ATE values that are numerically close. We use the following equation to test whether two floating point values, $a$ and $b$, are equivalent: 
$$ |a - b| <= (atol + rtol * |b|),$$
where $rtol$ is the relative tolerance parameter and $atol$ is the absolute tolerance parameter. Practically, we use the `isclose' function from the Numpy package\footnote{\url{https://numpy.org/doc/stable/reference/generated/numpy.isclose.html}} which uses the equation above and returns a boolean indicating whether $a$ and $b$ are equal within the given tolerance. We used the default values from Numpy, $rtol=1e-05$, $atol=1e-08$. We apply regrouping to the list of ATEs for precision and recall evaluation, but not for Wasserstein distance.

\end{document}